\documentclass{article}

\usepackage{arxiv}

\usepackage[utf8]{inputenc} 
\usepackage[T1]{fontenc}    
\usepackage{hyperref}       
\usepackage{url}            
\usepackage{booktabs}       
\usepackage{amsfonts}       
\usepackage{nicefrac}       
\usepackage{microtype}      
\usepackage{lipsum}		
\usepackage{graphicx}
\usepackage{natbib}
\usepackage{doi}

\title{MISAR: A Multimodal Instructional System with Augmented Reality}

\date{} 					

\author{ 
    Jing Bi\thanks{Equal contribution. \\ Approved for public release; distribution is unlimited. \\This work has been supported by the Defense Advance Research Projects Agency (DARPA) under Contract HR00112220003. The content of the information does not necessarily reflect the position of the Government, and no official endorsement should be inferred.} \\
    University of Rochester \\
    Rochester, NY \\
    \texttt{jing.bi@rochester.edu} \\
    \And
    Nguyen Manh Nguyen$^*$ \\
    University of Rochester \\
    Rochester, NY \\
    \texttt{nguyen.nguyen@rochester.edu} \\
    \And
    Ali Vosoughi$^*$ \\
    University of Rochester \\
    Rochester, NY \\
    \texttt{mvosough@ur.rochester.edu} \\
    \And
    Chenliang Xu \\
    University of Rochester \\
    Rochester, NY \\
    \texttt{chenliang.xu@rochester.edu} \\
}

\renewcommand{\headeright}{ICCV 2023 - AV4D}
\renewcommand{\undertitle}{accepted at ICCV 2023 - AV4D}
\renewcommand{\shorttitle}{MISAR}

\hypersetup{
pdftitle={MISAR: A Multimodal Instructional System with Augmented Reality},
pdfsubject={Computer Vision, Language, Augmented Reality, Egocentric Video},
}

\begin{document}
\maketitle

\begin{abstract}
Augmented reality (AR) requires the seamless integration of visual, auditory, and linguistic channels for optimized human-computer interaction. While auditory and visual inputs facilitate real-time and contextual user guidance, the potential of large language models (LLMs) in this landscape remains largely untapped. Our study introduces an innovative method harnessing LLMs to assimilate information from visual, auditory, and contextual modalities. Focusing on the unique challenge of task performance quantification in AR, we utilize egocentric video, speech, and context analysis. The integration of LLMs facilitates enhanced state estimation, marking a step towards more adaptive AR systems. Code, dataset, and demo will be available at \url{https://github.com/nguyennm1024/misar}.
\end{abstract}


\section{Introduction}

The effective design of augmented reality (AR) systems hinges on multimodal interfaces, blending visual, auditory, and linguistic elements to optimize human-computer interactions. When traditional text-based methods (e.g keyboard input) are not feasible or efficient for communication, auditory communication becomes vital, enabling real-time assistance even when visual information is limited. Similarly, visual data interpretation enhances AR system functionality by analyzing the user's environment, allowing for contextual support and proactive guidance. These auditory and visual capabilities together~\cite{Mo2022ACL} contribute to a more predictive and responsive AR system, thereby elevating the efficiency and accuracy of task completion.

The evolution of large language models (LLMs) has elevated language from a basic communication tool to a critical modality in computational reasoning and decision-making, integrating elements of context interpretation and inferential reasoning into the computational process \cite{zhao2023lavila, elizalde2023clap, li2022blip, girdhar2023imagebind}. While significant research has been dedicated to converting various sensory modalities into textual formats for applications such as video narration and complex dialog systems \cite{zhao2023lavila, vosoughi2023vqa, li2022auidovisualqa, elizalde2023clap}, these studies have yet to robustly address the intricate challenges specific to the augmented reality (AR) environment.

\begin{figure}
    \centering
    \includegraphics[width=0.99\linewidth]{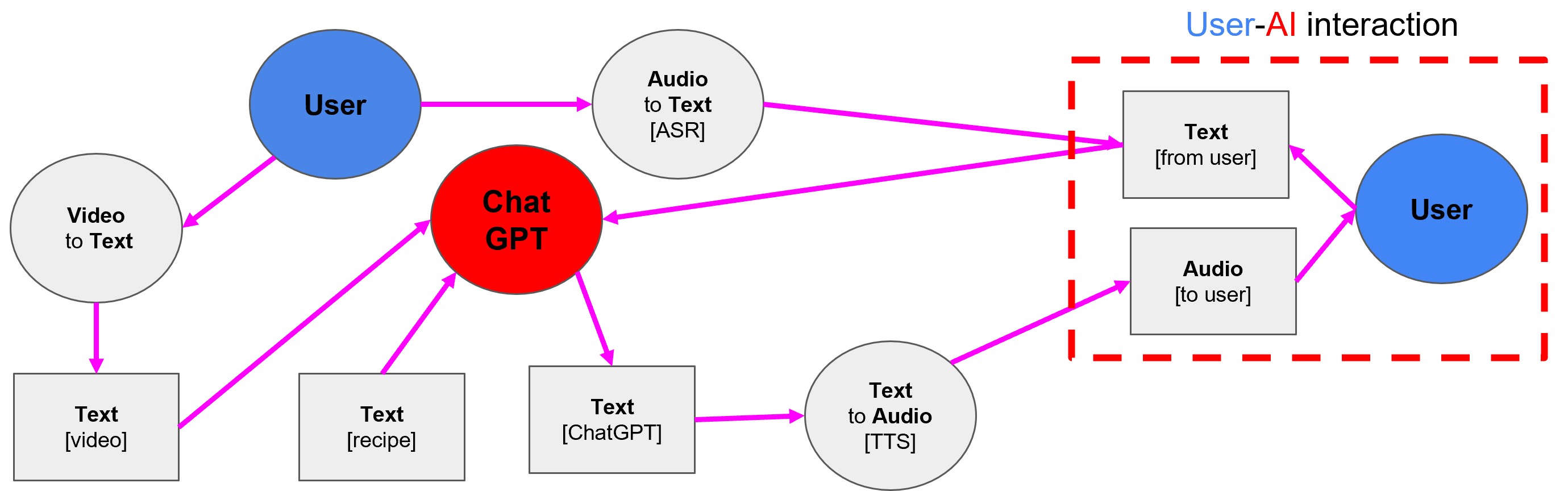}
    \caption{Architecture of the proposed multimodal integration model, highlighting GPT3.5-Turbo (\textit{aka} ChatGPT) as a central component. The model seamlessly integrates video-to-text, text-to-speech, audio-speech recognition, and GPT3.5-Turbo. However, errors may be induced during the video-to-text conversion process, and the model does not encompass understanding of environmental audio}
    \label{fig:conversational_chatpt}
\end{figure}

Building on previous research, we introduce a novel method that integrates multiple modalities—visual, auditory, and contextual—using language models for reasoning and human interaction \cite{zhao2023lavila, elizalde2023clap, li2022blip, girdhar2023imagebind}. Specifically, we address the unexplored problem of applying language models to AR for task performance quantification through egocentric video, speech, and context analysis. Our proposed architecture, illustrated in Fig. \ref{fig:conversational_chatpt}, positions GPT3.5-Turbo as the central component, aggregating information from diverse sources including visual cues, user and recipe inputs, and auditory data. These inputs inform GPT3.5-Turbo's contextually relevant responses and enable accurate state estimation of the user. The goal is to leverage the reasoning capabilities of language models to fill existing gaps in state estimation, moving towards a fully intelligent system.

In summary, our work makes two main contributions. First, we introduce a comprehensive approach that uses AI to assist people in working more accurately and efficiently. Second, we show that using Large Language Models (LLMs) to enhance video captions improves the quality of these descriptions, enabling effective scene understanding without training with domain specific data. These contributions significantly advance the field of intelligent computational systems.

\section{Related Works}
Our work encompasses multiple modalities, including speech, video, and natural language. In this section we highlight the related work to build the system that we aim.

\noindent \textbf{Large Language Models:} Large Language Models (LLMs), a remarkable advancement in artificial intelligence (AI), have instigated transformative changes across multiple disciplines, including computer vision~\cite{Zang2023ContextualOD,Nguyen2021DictionaryguidedST}, natural language processing~\cite{Nijkamp2022CodeGenAO,MacNeil2022GeneratingDC}, and speech processing~\cite{Rubenstein2023AudioPaLMAL,Chelba2012LargeSL}. Their implications extend to augmented and virtual reality (AR/VR), where they have the potential to serve as critical components. LLMs excel in language comprehension, reasoning, and response generation, rendering them effective decision-making cores in complex systems. In the proposed framework detailed in this paper, we employ GPT-3.5~\cite{Brown2020LanguageMA} from GPT3.5-Turbo to serve as the decision-making "brain" of the system. This highlights the pivotal role that LLMs, particularly GPT-3.5, can play in enhancing the capabilities of AR/VR systems.

\noindent \textbf{Video Captioning:} The task of generating natural language descriptions from video, encompassing both third-person and egocentric perspectives, has garnered significant attention within the research community. Recent studies have focused on learning video embeddings that are cross-attended with generative LLMs like GPT-2~\cite{Radford2019LanguageMA}, aiming to bridge the domain gap between textual and visual representations. In the present work, we employ a pretrained model known as LaViLa~\cite{zhao2023lavila}, for the specific task of video captioning.

\noindent\textbf{Speech Processing:} Speech is the most commonly used method of interaction among humans, so is that for the human-computer interaction also it can play a role. In this research we use google API for both speech to text (ASR) and text to speech (TTS) which was found in \cite{asr_google, tts_google}.

\section{Method}
Our system integrates ASR technology to transcribe verbal inputs into textual format, which is synergistically coupled with audiovisual data streams captured by the Hololens2 camera. This enables the system to acquire a comprehensive perspective of the user's visual field. The video footage procured from the camera is subsequently processed to generate summarizations and textual descriptions. LLMs serve as the computational epicenter of the system, fulfilling dual roles: they not only answer the user questions but also augment the fidelity of captioning in video descriptions, thereby enhancing the system's overall capability in scene understanding tasks.

\subsection{Recipe-based Instructional Dialog}
In this section, we will present the system's conversation feature. This feature is focused on the two most frequently asked questions of the support system, including asking about steps and asking about how to fix it when you get mistakes.

\noindent\textbf{Asking for Step Instructions:} As for the question type about the steps to perform a particular task, in our demo we experimented with the question types about what is the next step, what is the previous step. There is also a question form asking how each step is. These types of questions require the system to be able to understand the context of the conversation, understand the recipe, and understand the current state of the user.

\noindent\textbf{Asking for Fixing Mistakes} Errors are an inevitable aspect of human performance, particularly for individuals who lack prior experience in executing specific tasks. Consequently, the capacity for an assistive system to facilitate error correction becomes critically important. Providing solutions to rectify these mistakes necessitates that the system possess both generalized knowledge across multiple domains as well as contextual understanding specific to the user's current situation. This dual requirement underscores the need for a nuanced and adaptive system capable of offering effective support for error mitigation.

\subsection{Enhancing Video Captions}

\begin{figure}
    \centering
    \includegraphics[width=0.99\linewidth]{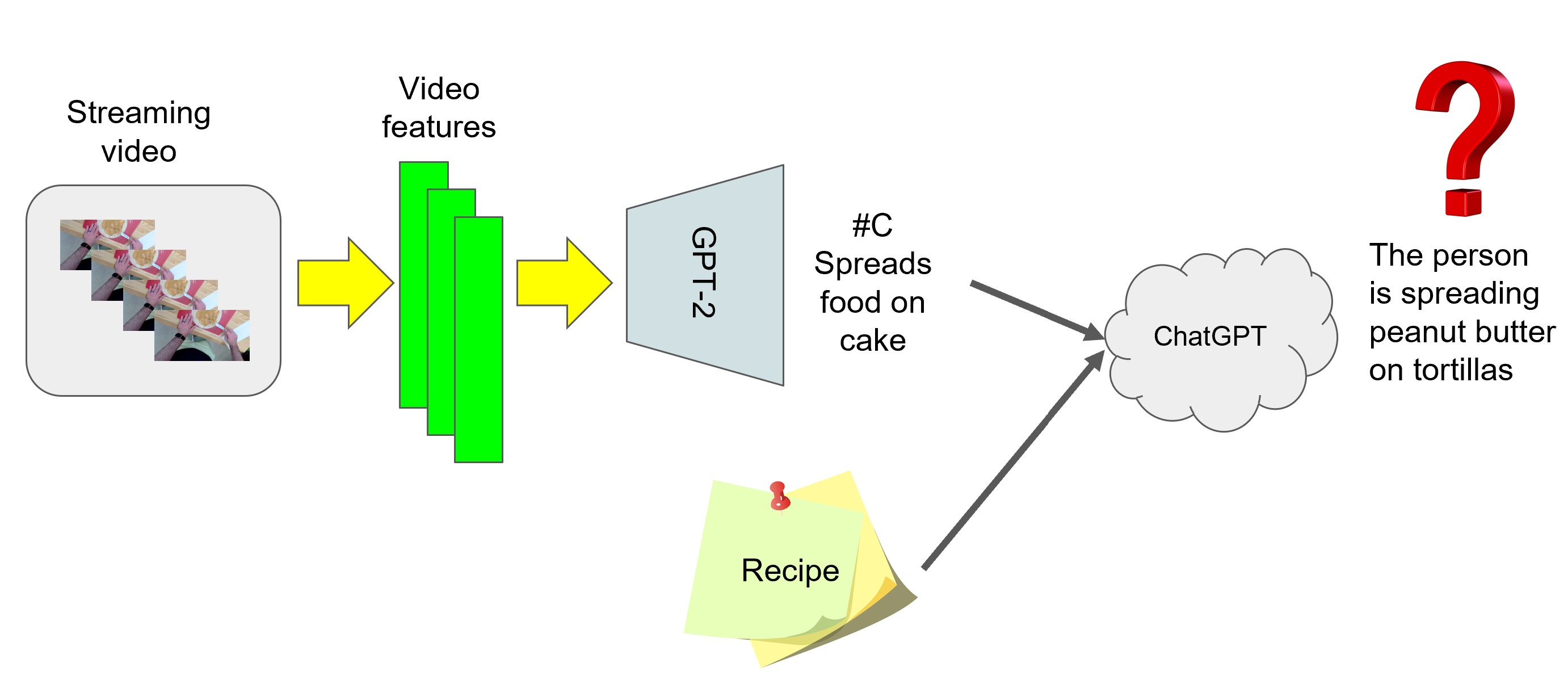}
    \caption{Architecture of the proposed model showcasing the integration of GPT3.5-Turbo (ChatGPT) into the workflow. By incorporating GPT3.5-Turbo, we enhance the quality of generated text descriptions and establish a seamless means of communication between instructional input, users, and the video text.}
    \label{fig:chatgpt_for_improving}
\end{figure}

By integrating GPT3.5-Turbo into the workflow, we can improve the generated text descriptions and establish an effective means of communication between instructional input, users, and the video text. The architecture of the proposed model is shown in Fig. \ref{fig:chatgpt_for_improving}.

\begin{figure}
    \centering
    \includegraphics[width=0.99\linewidth]{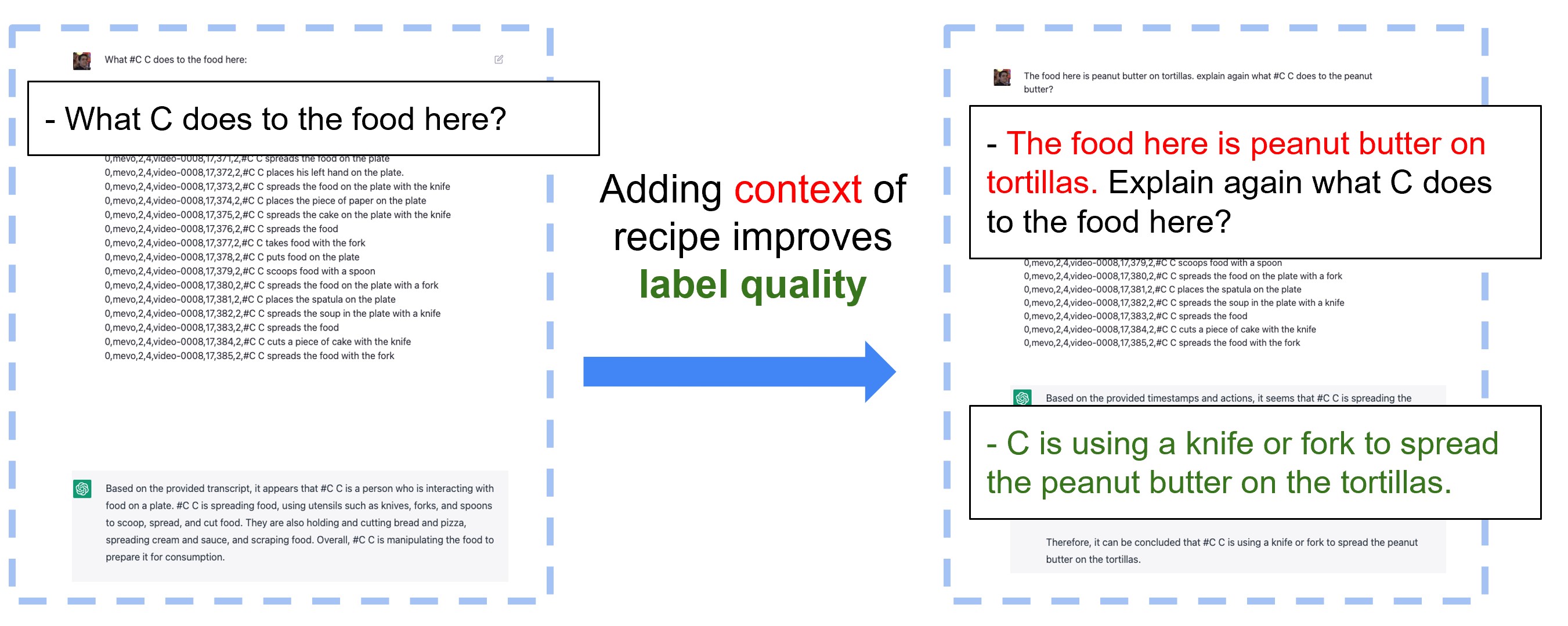}
    \caption{The impact of leveraging larger models such as GPT3.5-Turbo on enhancing the quality of textual labels within the given context. Leveraging larger LLMs and inclusion of contextual information such as recipes or instructions significantly improve the accuracy and richness of generated textual labels, leading to a better understanding and interpretation of the content."}
    \label{fig:enhancing_labels_chatgpt}
\end{figure}

The integration of GPT3.5-Turbo as a central communication point allows for a seamless interaction between users and the video content. Users can provide instructions or ask questions related to specific video segments, and GPT3.5-Turbo responds with informative and contextually relevant explanations. This dynamic interaction not only improves the accuracy and richness of the generated video-to-text labels but also empowers users to actively engage with the content and obtain valuable insights. As shown in Fig. \ref{fig:enhancing_labels_chatgpt}, leveraging larger models like GPT3.5-Turbo allows for substantial improvements in the quality of textual labels within the given context. The application of these advanced models enhances the accuracy and richness of generated textual labels, ultimately enhancing the overall understanding and interpretation of the content.

\subsection{An User-friendly Interactive System}
In the comprehensive system we propose, users can effortlessly engage via AR glasses, facilitating a streamlined human-computer interaction predominantly through speech. The interactive experience is designed to emulate collaboration with a seasoned expert, thereby elevating the user's overall task performance. Moreover, the system serves as a cost-effective alternative to traditional methods of advanced human training, obviating the need for the hiring of specialized instructors. This approach offers a scalable solution for enhancing human performance while reducing associated training costs.

\section{How to scale up with multimodal LLMs guidance? }

\begin{figure}
    \centering
    \includegraphics[width=0.99\linewidth]{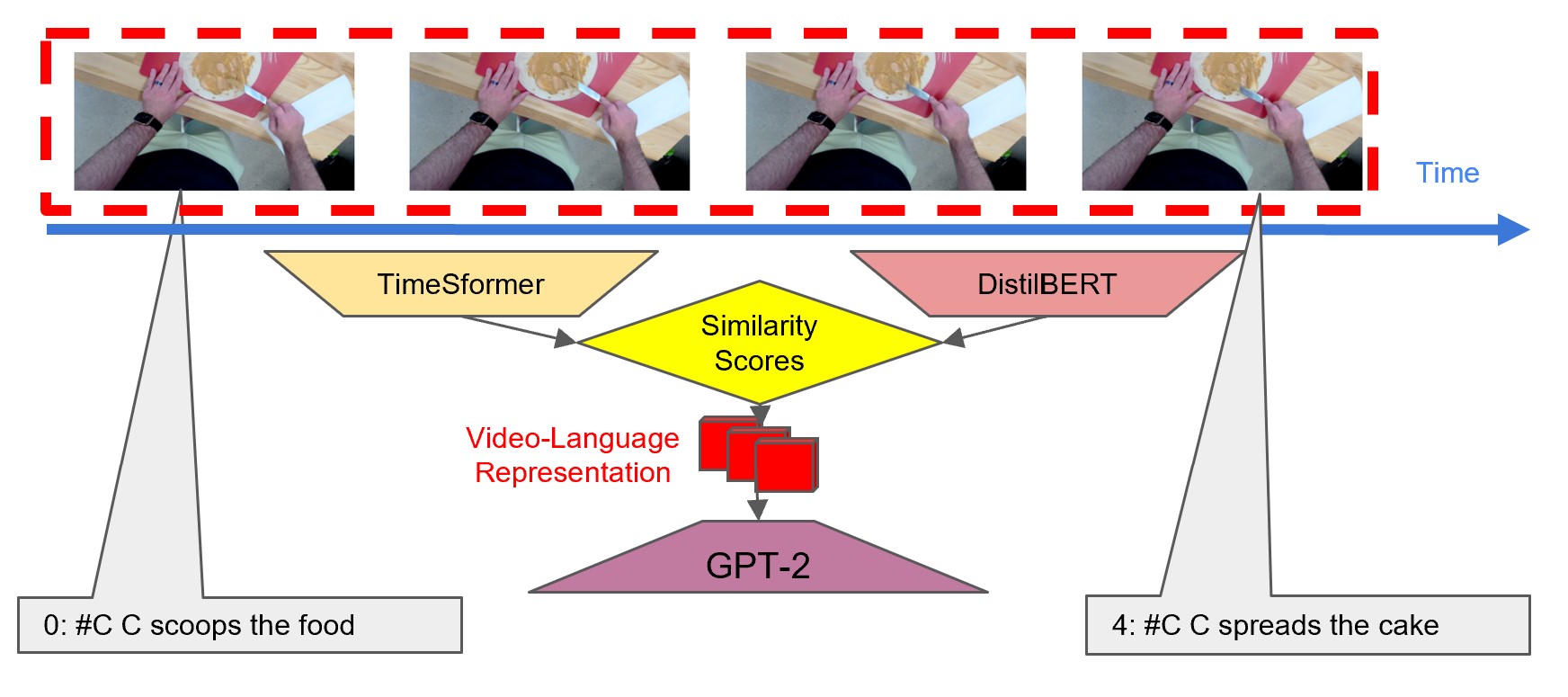}
    \caption{Converting visual content to descriptive text with Video-Conditioned Text Generation using LLMs.}
    \label{fig:gpt2_video_to_text}
\end{figure}

In our video-to-text pipeline, we utilized the LaViLa model \cite{zhao2023lavila}, as depicted in Fig.~\ref{fig:gpt2_video_to_text}. Pre-trained on the Ego4D dataset~\cite{Grauman2021Ego4DAT}, the model is designed to transfer this expertise effectively to Perceptually-enabled Task Guidance (PTG) data, thereby achieving accurate and coherent text generation from video inputs.

\subsection{Scaling the Natural Language Labeling of Videos}
\begin{figure}
    \centering
    \includegraphics[width=0.99\linewidth]{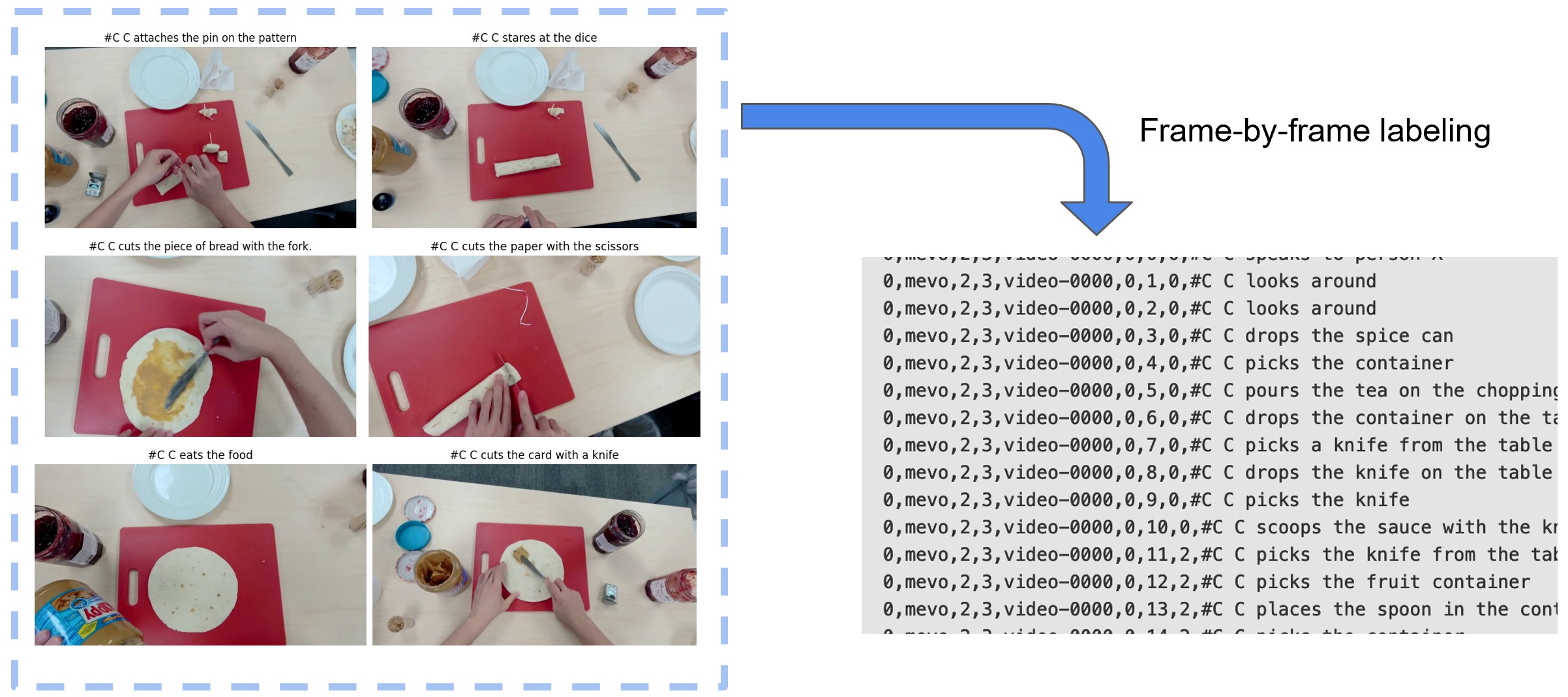}
    \caption{Automating frame-by-frame labeling and annotation of video files through video-to-text conversion models. This scalable and versatile approach enables efficient natural language labeling across diverse domains. However, it is important to note that the generated text may contain noise, lack crucial details and environmental dynamics, or even produce unrelated annotations.}
    \label{fig:scaling_labels_using_llms}
\end{figure}

We explore the application of video-to-text conversion as a means of automating the frame-by-frame labeling and annotation process for video files. In this approach, we leverage video Transformers and GPT-2~\cite{Radford2019LanguageMA} decoder with cross-modal attention for video understanding, visual-linguistic attention, and natural language generation to streamline the arduous task of labeling videos with textual descriptions. An example of the automated annotation of the videos is shown in Fig.~\ref{fig:scaling_labels_using_llms}. By leveraging pre-trained models, video-to-text conversion enables efficient  analysis of the visual content, resulting in the generation of descriptive labels for each frame in the video. Moreover, with its ability to handle large-scale video datasets, video-to-text conversion presents a scalable solution for effectively scaling up the natural language labeling of videos, opening doors to more extensive applications in various domains. Despite the benefits, having language decoders such as GPT-2 makes it vulnerable to noisy label generation, which sometimes is not related to the scene, lacks sufficient information, attends to wrong or less important details, or is entirely unrelated/wrong text. This encourages us to use an even larger LLM to help polish the results of the GPT-2 here.

\subsection{Multimodal LLMs Guidance and User-AI Interaction}
The proposed model offers a comprehensive integration of various modalities, including video-to-text, text-to-speech, audio-speech recognition, and GPT3.5-Turbo. This multimodal approach ensures that all forms of input, regardless of their original format, are converted into text representations. By unifying the data into a common textual format, the model facilitates seamless integration and interaction across different modalities.

By combining the capabilities of video-to-text, text-to-speech, audio-speech recognition, and GPT3.5-Turbo, the proposed model enables a conversational and interactive experience. Users can provide instructions, queries, or engage in natural language conversations with the system, which responds accordingly based on the fused textual representations. This holistic approach enhances the model's ability to understand and interpret diverse input types, enabling more comprehensive and effective communication between users and the system.

\section{Experiments}
\noindent \textbf{Data:} We collated in-house data on an infrequently documented pinwheel recipe. With the appropriate legal clearances, we utilized a helmet with video and audio functionalities in our lab to record participants making the recipe, ensuring hands were also captured (unlike HoloLens2, which does not record hands, leading us to exclude the use of HL2SS). The resulting videos have varying duration, and they come annotated with the 13 steps of the method. A subset of this data is shared on our GitHub repository.

\noindent \textbf{Video Conversion with LaViLa:} For each frame, we produce a sentence, given the video's frame rate of 30 fps. Considering that the quickest human response and action takes around 250 ms, generating captions for every 8 frames aligns with real-time performance needs in physical augmented reality. This frequency matches the swiftest observable human movement.

\noindent \textbf{Results and Discussion:} In Tab.~\ref{tab:similarity}, the results show that the text after being polished with LLMs was significantly more relevant to the context under consideration. From the trained video to text model with general data, we have made the generated text more contextual, thereby opening the potential for reducing the domain gap between pretrained models and domain specific data without finetuning is required. From there, it shows that our proposed method can be easily adapted to many different domains easily. Tab.~\ref{tab:step_wise_similarity} illustrates the impact of recipe types on the quality of generated captions. The data reveal that medium-length recipe descriptions yield captions most closely aligned with the intended context. This outcome is logical; brief descriptions lack sufficient information for LLMs to accurately capture context, while overly extensive descriptions introduce noise that compromises caption quality.
\begin{figure}
    \centering
    \includegraphics[width=0.9\linewidth]{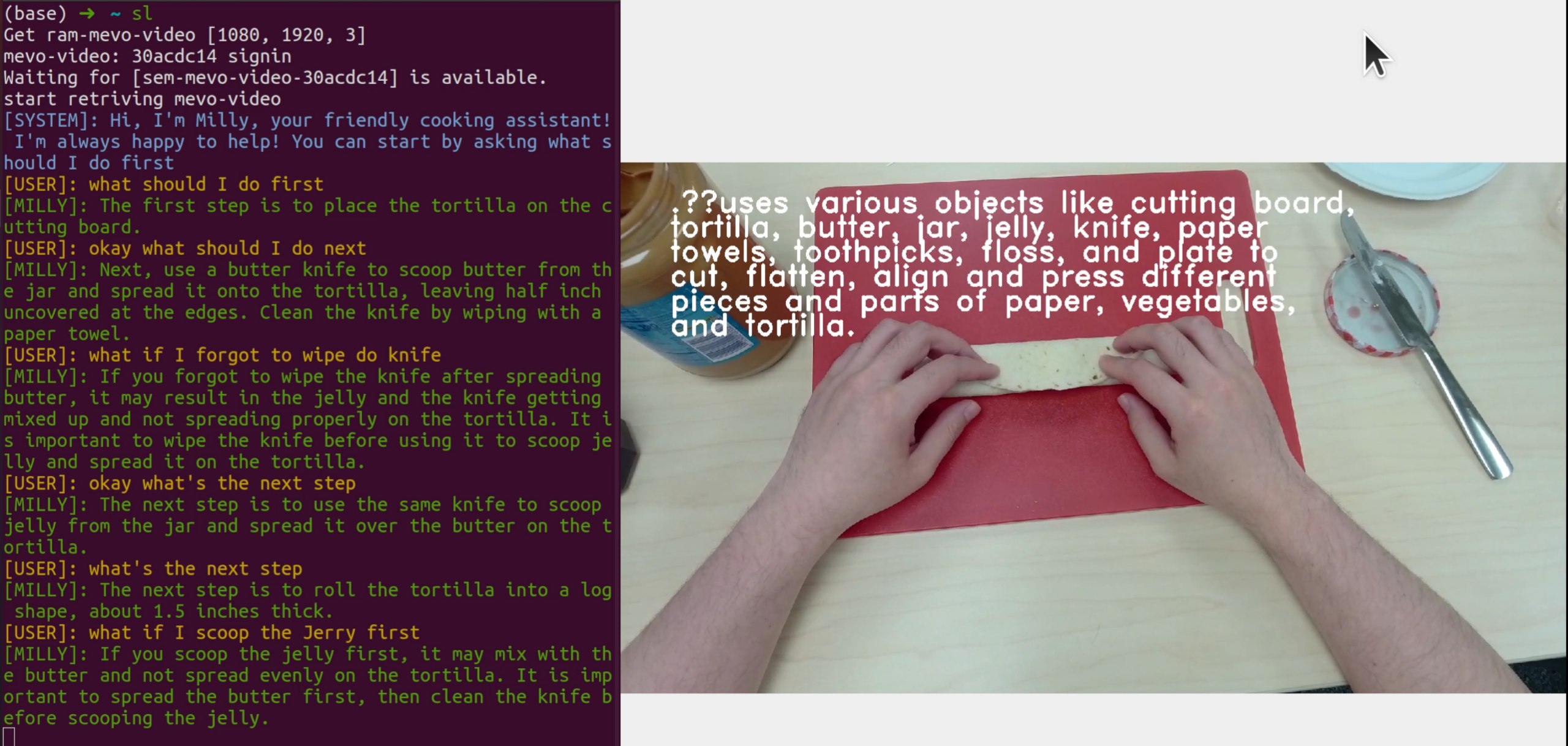}
    \caption{Screenshot of the demo showcasing the dialog system. Full video is available in our GitHub repository.}
    \label{fig:demo_screenshot}
\end{figure}
We demonstrate the efficacy of our dialog system through a live demo. The system is user-friendly and capable of providing immediate responses to user inquiries. During the evaluation phase, the system was subjected to a variety of question types, including those that describe requisite actions for task completion as well as troubleshooting procedures for errors. A screenshot of the demo is shown in Fig. \ref{fig:demo_screenshot}. The demonstration video can be accessed at our Github.

\begin{table}
\centering
\begin{tabular}{|c|c|c|}
\hline
\textbf{Reference recipe}& \textbf{LaViLa \cite{zhao2023lavila}} & \textbf{Ours} \\
\hline
{2-Words Text} & 0.817 & \textbf{0.845} \\
\hline
{Full sentence} & 0.841 & \textbf{0.854 }\\
\hline
{Descriptive sentences} & 0.834 & \textbf{0.848} \\
\hline
\end{tabular}
\caption{The numbers indicates the ability of the proposed model to classify the step based on natural language descriptions. The LaViLa method has been compared to our method. The recipe texts used to compute similarity are referred to at three different levels: \textit{2-Words text}, \textit{full sentence}, and \textit{descriptive sentences}. Examples are \textit{"spread butter"}, \textit{"Evenly spread butter on the tortilla"}, and \textit{"Gently spread the scooped butter evenly over the entire surface of the tortilla, leaving a small margin around the edges to allow for spreading when you roll it"}
, respectively.}
\label{tab:similarity}
\end{table}

\begin{table}
\centering
\resizebox{0.95\textwidth}{!}{
\begin{tabular}{|c|*{13}{c|}}
\hline
\bfseries Step & \bfseries 0 & \bfseries 1 & \bfseries 2 & \bfseries 3 & \bfseries 4 & \bfseries 5 & \bfseries 6 & \bfseries 7 & \bfseries 8 & \bfseries 9 & \bfseries 10 & \bfseries 11 & \bfseries 12 \\
\hline
\#Samples & \textbf{11} & \textbf{16} & \textbf{773} & \textbf{152} & \textbf{523} & \textbf{153} & \textbf{387} & \textbf{466} & \textbf{407} & \textbf{338} & \textbf{355} & \textbf{1752} & \textbf{338} \\
\hline
LaViLa - short & 0.830 & 0.810 & 0.825 & 0.810 & 0.849 & 0.823 & 0.822 & 0.802 & 0.838 & 0.807 & 0.797 & 0.805 & 0.839 \\
\hline
LaViLa - medium & 0.823 & 0.840 & 0.849 & 0.837 & 0.842 & 0.840 & 0.847 & 0.845 & 0.847 & 0.840 & 0.835 & 0.835 & 0.844 \\
\hline
LaViLa - long & 0.822 & 0.830 & 0.834 & 0.833 & 0.845 & 0.834 & 0.838 & 0.838 & 0.835 & 0.834 & 0.836 & 0.831 & 0.827 \\
\hline
Ours - short & 0.843 & 0.863 & 0.846 & 0.829 & 0.849 & 0.841 & 0.841 & 0.842 & 0.850 & 0.845 & 0.811 & 0.846 & \textbf{0.873} \\
\hline
Ours - medium & \textbf{0.845} & \textbf{0.865} & \textbf{0.853} & \textbf{0.858} & \textbf{0.847} & \textbf{0.858} & \textbf{0.849} & \textbf{0.862} & \textbf{0.854} & \textbf{0.848} & \textbf{0.840} & \textbf{0.858} & 0.859 \\
\hline
Ours - long & 0.842 & 0.847 & 0.846 & 0.850 & 0.847 & 0.855 & \textbf{0.847} & 0.847 & 0.851 & 0.844 & 0.842 & 0.851 & 0.841 \\
\hline
\end{tabular}
}
\caption{The values represent step-wise similarity scores between the generated narratives and the reference recipe descriptions. The table specifies the number of samples for each step, all originating from a single video. Here, "short" corresponds to a 2-word text, "medium" is a full sentence, and "long" refers to detailed sentences as seen in \ref{tab:similarity}. }
\label{tab:step_wise_similarity}
\end{table}

\section{Conclusion}
In summary, this paper introduces a system designed to enhance both human's accuracy and efficiency in task performance. Additionally, we advocate for the incorporation of LLMs to improve the quality of video captions. This enhancement opens the way for more effective text-based scene understanding. Our methodology's effectiveness is substantiated through both experimental outcomes and system demo.

\bibliographystyle{unsrtnat}
\bibliography{references}  

\begin{thebibliography}{18}
\providecommand{\natexlab}[1]{#1}
\providecommand{\url}[1]{\texttt{#1}}
\expandafter\ifx\csname urlstyle\endcsname\relax
  \providecommand{\doi}[1]{doi: #1}\else
  \providecommand{\doi}{doi: \begingroup \urlstyle{rm}\Url}\fi

\bibitem[Mo and Morgado(2022)]{Mo2022ACL}
Shentong Mo and Pedro Morgado.
\newblock A closer look at weakly-supervised audio-visual source localization.
\newblock \emph{ArXiv}, abs/2209.09634, 2022.
\newblock URL \url{https://api.semanticscholar.org/CorpusID:252383734}.

\bibitem[Zhao et~al.(2023)Zhao, Misra, Kr{\"a}henb{\"u}hl, and Girdhar]{zhao2023lavila}
Yue Zhao, Ishan Misra, Philipp Kr{\"a}henb{\"u}hl, and Rohit Girdhar.
\newblock Learning video representations from large language models.
\newblock In \emph{Proceedings of the IEEE/CVF Conference on Computer Vision and Pattern Recognition}, pages 6586--6597, 2023.

\bibitem[Elizalde et~al.(2023)Elizalde, Deshmukh, Al~Ismail, and Wang]{elizalde2023clap}
Benjamin Elizalde, Soham Deshmukh, Mahmoud Al~Ismail, and Huaming Wang.
\newblock Clap learning audio concepts from natural language supervision.
\newblock In \emph{ICASSP 2023-2023 IEEE International Conference on Acoustics, Speech and Signal Processing (ICASSP)}, pages 1--5. IEEE, 2023.

\bibitem[Li et~al.(2022{\natexlab{a}})Li, Li, Xiong, and Hoi]{li2022blip}
Junnan Li, Dongxu Li, Caiming Xiong, and Steven Hoi.
\newblock Blip: Bootstrapping language-image pre-training for unified vision-language understanding and generation.
\newblock In \emph{International Conference on Machine Learning}, pages 12888--12900. PMLR, 2022{\natexlab{a}}.

\bibitem[Girdhar et~al.(2023)Girdhar, El-Nouby, Liu, Singh, Alwala, Joulin, and Misra]{girdhar2023imagebind}
Rohit Girdhar, Alaaeldin El-Nouby, Zhuang Liu, Mannat Singh, Kalyan~Vasudev Alwala, Armand Joulin, and Ishan Misra.
\newblock Imagebind: One embedding space to bind them all.
\newblock In \emph{Proceedings of the IEEE/CVF Conference on Computer Vision and Pattern Recognition}, pages 15180--15190, 2023.

\bibitem[Vosoughi et~al.(2023)Vosoughi, Deng, Zhang, Tian, Xu, and Luo]{vosoughi2023vqa}
Ali Vosoughi, Shijian Deng, Songyang Zhang, Yapeng Tian, Chenliang Xu, and Jiebo Luo.
\newblock Unveiling cross modality bias in visual question answering: A causal view with possible worlds vqa.
\newblock \emph{arXiv preprint arXiv:2305.19664}, 2023.

\bibitem[Li et~al.(2022{\natexlab{b}})Li, Wei, Tian, Xu, Wen, and Hu]{li2022auidovisualqa}
Guangyao Li, Yake Wei, Yapeng Tian, Chenliang Xu, Ji-Rong Wen, and Di~Hu.
\newblock Learning to answer questions in dynamic audio-visual scenarios.
\newblock In \emph{Proceedings of the IEEE/CVF Conference on Computer Vision and Pattern Recognition}, pages 19108--19118, 2022{\natexlab{b}}.

\bibitem[Zang et~al.(2023)Zang, Li, Han, Zhou, and Loy]{Zang2023ContextualOD}
Yuhang Zang, Wei Li, Jun Han, Kaiyang Zhou, and Chen~Change Loy.
\newblock Contextual object detection with multimodal large language models.
\newblock \emph{ArXiv}, abs/2305.18279, 2023.
\newblock URL \url{https://api.semanticscholar.org/CorpusID:258959011}.

\bibitem[Nguyen et~al.(2021)Nguyen, Nguyen, Tran, Tran, Ngo, Nguyen, and Hoai]{Nguyen2021DictionaryguidedST}
Nguyen~Manh Nguyen, Thu Nguyen, Vinh Tran, Minh-Triet Tran, Thanh~Duc Ngo, Thien~Huu Nguyen, and Minh Hoai.
\newblock Dictionary-guided scene text recognition.
\newblock \emph{2021 IEEE/CVF Conference on Computer Vision and Pattern Recognition (CVPR)}, pages 7379--7388, 2021.
\newblock URL \url{https://api.semanticscholar.org/CorpusID:235341891}.

\bibitem[Nijkamp et~al.(2022)Nijkamp, Pang, Hayashi, Tu, Wang, Zhou, Savarese, and Xiong]{Nijkamp2022CodeGenAO}
Erik Nijkamp, Bo~Pang, Hiroaki Hayashi, Lifu Tu, Haiquan Wang, Yingbo Zhou, Silvio Savarese, and Caiming Xiong.
\newblock Codegen: An open large language model for code with multi-turn program synthesis.
\newblock In \emph{International Conference on Learning Representations}, 2022.
\newblock URL \url{https://api.semanticscholar.org/CorpusID:252668917}.

\bibitem[MacNeil et~al.(2022)MacNeil, Tran, Mogil, Bernstein, Ross, and Huang]{MacNeil2022GeneratingDC}
Stephen MacNeil, Andrew Tran, Dan Mogil, Seth Bernstein, Erin Ross, and Ziheng Huang.
\newblock Generating diverse code explanations using the gpt-3 large language model.
\newblock \emph{Proceedings of the 2022 ACM Conference on International Computing Education Research - Volume 2}, 2022.
\newblock URL \url{https://api.semanticscholar.org/CorpusID:251322329}.

\bibitem[Rubenstein et~al.(2023)Rubenstein, Asawaroengchai, Nguyen, Bapna, Borsos, de~Chaumont~Quitry, Chen, Badawy, Han, Kharitonov, Muckenhirn, Padfield, Qin, Rozenberg, Sainath, Schalkwyk, Sharifi, Tadmor, Ramanovich, Tagliasacchi, Tudor, Velimirovi'c, Vincent, Yu, Wang, Zayats, Zeghidour, Zhang, Zhang, Zilka, and Frank]{Rubenstein2023AudioPaLMAL}
Paul~K. Rubenstein, Chulayuth Asawaroengchai, Duc~Dung Nguyen, Ankur Bapna, Zal{\'a}n Borsos, F{\'e}lix de~Chaumont~Quitry, Peter Chen, Dalia~El Badawy, Wei Han, Eugene Kharitonov, Hannah Muckenhirn, Dirk~Ryan Padfield, James Qin, Daniel Rozenberg, Tara~N. Sainath, Johan Schalkwyk, Matthew Sharifi, Michelle~D. Tadmor, Ramanovich, Marco Tagliasacchi, Alexandru Tudor, Mihajlo Velimirovi'c, Damien Vincent, Jiahui Yu, Yongqiang Wang, Victoria Zayats, Neil Zeghidour, Yu~Zhang, Zhishuai Zhang, Luk{\'a}s Zilka, and Christian~Havn{\o} Frank.
\newblock Audiopalm: A large language model that can speak and listen.
\newblock \emph{ArXiv}, abs/2306.12925, 2023.
\newblock URL \url{https://api.semanticscholar.org/CorpusID:259224345}.

\bibitem[Chelba et~al.(2012)Chelba, Bikel, Shugrina, Nguyen, and Kumar]{Chelba2012LargeSL}
Ciprian Chelba, Daniel~M. Bikel, Maria Shugrina, Patrick Nguyen, and Shankar Kumar.
\newblock Large scale language modeling in automatic speech recognition.
\newblock \emph{ArXiv}, abs/1210.8440, 2012.
\newblock URL \url{https://api.semanticscholar.org/CorpusID:15170494}.

\bibitem[Brown et~al.(2020)Brown, Mann, Ryder, Subbiah, Kaplan, Dhariwal, Neelakantan, Shyam, Sastry, Askell, Agarwal, Herbert-Voss, Krueger, Henighan, Child, Ramesh, Ziegler, Wu, Winter, Hesse, Chen, Sigler, Litwin, Gray, Chess, Clark, Berner, McCandlish, Radford, Sutskever, and Amodei]{Brown2020LanguageMA}
Tom~B. Brown, Benjamin Mann, Nick Ryder, Melanie Subbiah, Jared Kaplan, Prafulla Dhariwal, Arvind Neelakantan, Pranav Shyam, Girish Sastry, Amanda Askell, Sandhini Agarwal, Ariel Herbert-Voss, Gretchen Krueger, T.~J. Henighan, Rewon Child, Aditya Ramesh, Daniel~M. Ziegler, Jeff Wu, Clemens Winter, Christopher Hesse, Mark Chen, Eric Sigler, Mateusz Litwin, Scott Gray, Benjamin Chess, Jack Clark, Christopher Berner, Sam McCandlish, Alec Radford, Ilya Sutskever, and Dario Amodei.
\newblock Language models are few-shot learners.
\newblock \emph{ArXiv}, abs/2005.14165, 2020.
\newblock URL \url{https://api.semanticscholar.org/CorpusID:218971783}.

\bibitem[Radford et~al.(2019)Radford, Wu, Child, Luan, Amodei, and Sutskever]{Radford2019LanguageMA}
Alec Radford, Jeff Wu, Rewon Child, David Luan, Dario Amodei, and Ilya Sutskever.
\newblock Language models are unsupervised multitask learners.
\newblock 2019.
\newblock URL \url{https://api.semanticscholar.org/CorpusID:160025533}.

\bibitem[asr()]{asr_google}
https://cloud.google.com/speech-to-text.

\bibitem[tts()]{tts_google}
https://cloud.google.com/text-to-speech.

\bibitem[Grauman et~al.(2021)Grauman, Westbury, Byrne, Chavis, Furnari, Girdhar, Hamburger, Jiang, Liu, Liu, Martin, Nagarajan, Radosavovic, Ramakrishnan, Ryan, Sharma, Wray, Xu, Xu, Zhao, Bansal, Batra, Cartillier, Crane, Do, Doulaty, Erapalli, Feichtenhofer, Fragomeni, Fu, Fuegen, Gebreselasie, Gonz{\'a}lez, Hillis, Huang, Huang, Jia, Khoo, Kol{\'a}r, Kottur, Kumar, Landini, Li, Li, Li, Mangalam, Modhugu, Munro, Murrell, Nishiyasu, Price, Puentes, Ramazanova, Sari, Somasundaram, Southerland, Sugano, Tao, Vo, Wang, Wu, Yagi, Zhu, Arbel{\'a}ez, Crandall, Damen, Farinella, Ghanem, Ithapu, Jawahar, Joo, Kitani, Li, Newcombe, Oliva, Park, Rehg, Sato, Shi, Shou, Torralba, Torresani, Yan, and Malik]{Grauman2021Ego4DAT}
Kristen Grauman, Andrew Westbury, Eugene Byrne, Zachary~Q. Chavis, Antonino Furnari, Rohit Girdhar, Jackson Hamburger, Hao Jiang, Miao Liu, Xingyu Liu, Miguel Martin, Tushar Nagarajan, Ilija Radosavovic, Santhosh~K. Ramakrishnan, Fiona Ryan, Jayant Sharma, Michael Wray, Mengmeng Xu, Eric~Z. Xu, Chen Zhao, Siddhant Bansal, Dhruv Batra, Vincent Cartillier, Sean Crane, Tien Do, Morrie Doulaty, Akshay Erapalli, Christoph Feichtenhofer, Adriano Fragomeni, Qichen Fu, Christian Fuegen, Abrham Gebreselasie, Cristina Gonz{\'a}lez, James~M. Hillis, Xuhua Huang, Yifei Huang, Wenqi Jia, Weslie Khoo, J{\'a}chym Kol{\'a}r, Satwik Kottur, Anurag Kumar, Federico Landini, Chao Li, Yanghao Li, Zhenqiang Li, Karttikeya Mangalam, Raghava Modhugu, Jonathan Munro, Tullie Murrell, Takumi Nishiyasu, Will Price, Paola~Ruiz Puentes, Merey Ramazanova, Leda Sari, Kiran~K. Somasundaram, Audrey Southerland, Yusuke Sugano, Ruijie Tao, Minh Vo, Yuchen Wang, Xindi Wu, Takuma Yagi, Yunyi Zhu, Pablo Arbel{\'a}ez, David~J. Crandall, Dima Damen,
  Giovanni~Maria Farinella, Bernard Ghanem, Vamsi~Krishna Ithapu, C.~V. Jawahar, Hanbyul Joo, Kris Kitani, Haizhou Li, Richard~A. Newcombe, Aude Oliva, Hyun~Soo Park, James~M. Rehg, Yoichi Sato, Jianbo Shi, Mike~Zheng Shou, Antonio Torralba, Lorenzo Torresani, Mingfei Yan, and Jitendra Malik.
\newblock Ego4d: Around the world in 3,000 hours of egocentric video.
\newblock \emph{2022 IEEE/CVF Conference on Computer Vision and Pattern Recognition (CVPR)}, pages 18973--18990, 2021.
\newblock URL \url{https://api.semanticscholar.org/CorpusID:238856888}.

\end{thebibliography}






\end{document}